\documentclass[10pt,twocolumn,letterpaper]{article}

\usepackage[pagenumbers]{cvpr} 

\usepackage[dvipsnames]{xcolor}
\usepackage{lipsum}
\usepackage[framemethod=TikZ]{mdframed}
\usepackage{siunitx}

\newcommand{\benchmark}{D\textsubscript{3}}

\definecolor{cvprblue}{rgb}{0.21,0.49,0.74}
\usepackage[pagebackref,breaklinks,colorlinks,citecolor=cvprblue]{hyperref}

\def\paperID{*****} 
\def\confName{ECCVW: EVAL-FoMo}
\def\confYear{2024}

\title{Detect, Describe, Discriminate: Moving Beyond VQA for MLLM Evaluation}

\author{Manu Gaur \hspace{6mm}
Darshan Singh S \hspace{6mm}
Makarand Tapaswi \\
CVIT, IIIT Hyderabad \\
{\small \url{https://katha-ai.github.io/projects/detect-describe-discriminate/}}
}

\begin{document}

\twocolumn[{%
\renewcommand\twocolumn[1][]{#1}%
\maketitle
\centering
\vspace{-2mm}
\includegraphics[width=0.97\linewidth]{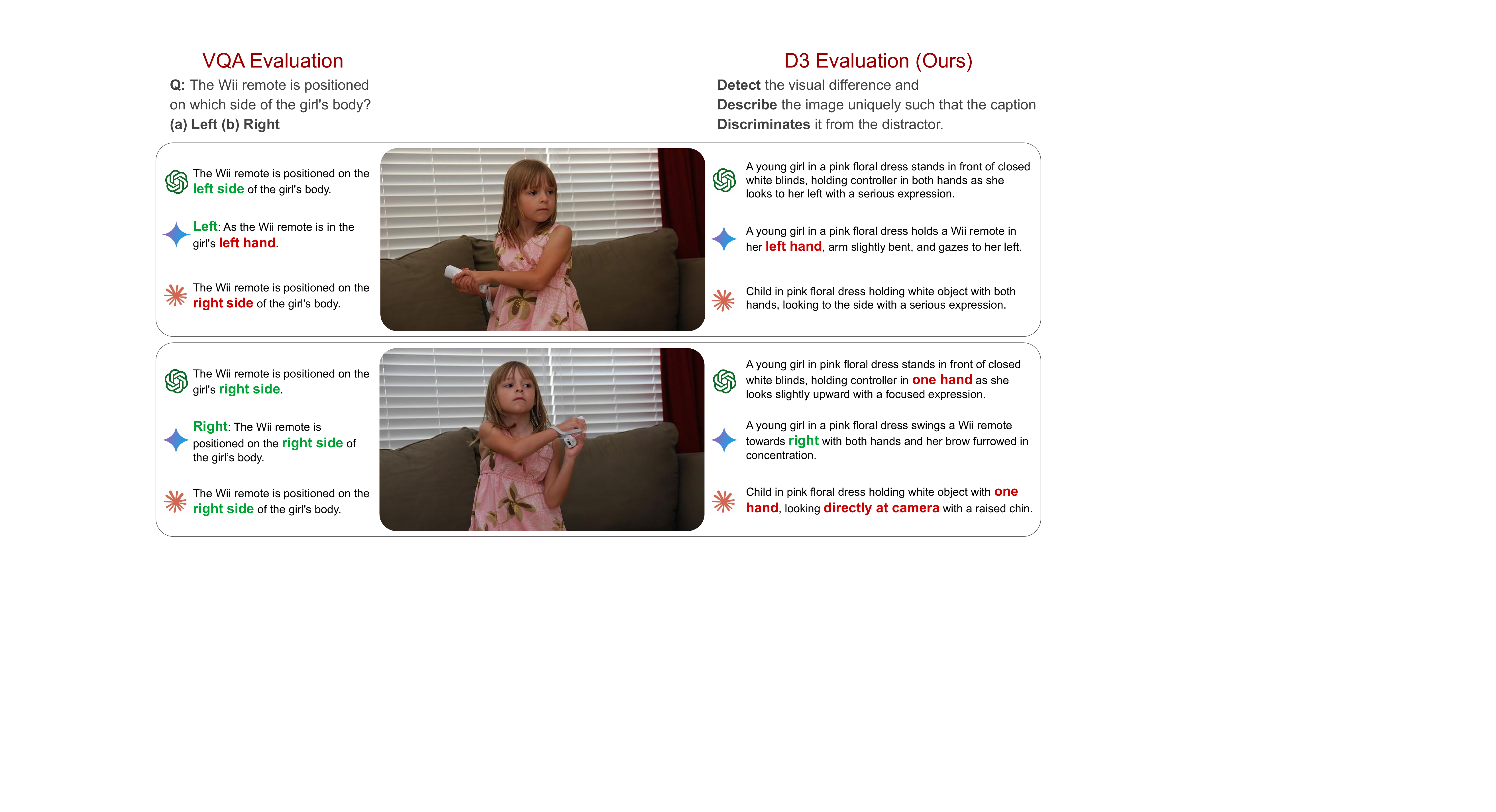}
\vspace{-3mm}
\captionof{figure}{
When prompted with a question and/or multiple choices (\textit{VQA evaluation}), MLLMs show middling performance on identifying fine-grained differences between an image pair.
Harder still, is when MLLMs need to independently detect and describe such differences (\textit{our evaluation}).
Our work finds that state-of-the-art MLLMs struggle to discern fine-grained difference with our detect-describe-discriminate evaluation framework, with open-source MLLMs failing to outperform random guess.
The text highlighted in \textbf{\textcolor{ForestGreen}{green}} represents the fine-grained differences captured by the MLLMs, while that marked in \textbf{\textcolor{BrickRed}{red}} represents erroneous descriptions (hallucinations).
Results are presented for GPT-4o, Gemini-1.5-Pro, and Claude-Sonnet-3.5.
}
\label{fig:vqa_teaser}
\vspace{6mm}
}]

\begin{abstract}
\label{sec:abstract}
\vspace{-2mm}
Visual Question Answering (VQA) with multiple choice questions enables a vision-centric evaluation of Multimodal Large Language Models (MLLMs).
Although it reliably checks the 
existence of specific visual abilities, 
it is easier for the model to select an answer from multiple choices (VQA evaluation) than to generate the answer itself.
In this work, we offer a novel perspective: we evaluate how well an MLLM understands a specific visual concept by its ability to uniquely describe two extremely similar images that differ only in the targeted visual concept.
Specifically, we assess the ability of MLLMs to capture specific points of visual differences using self-retrieval~\cite{liu2018selfretrieval}, \ie by retrieving the target image using its generated caption against the other image in the pair serving as the distractor.
We curate 247 highly similar image pairs as part of the \textbf{\benchmark} benchmark.
For each image pair, the model is prompted to:
(1)~\textbf{Detect} a specific visual difference, and
(2)~\textbf{Describe} the target image uniquely such that it
(3)~\textbf{Discriminate}s the target image from the distractor.
Self-retrieval within \benchmark{} enables whitebox evaluation across six different visual patterns, revealing that current models struggle to independently discern fine-grained visual differences, with open-source models failing to outperform random guess.

\end{abstract}
\vspace{-2mm}

\section{Introduction}
\label{sec:intro}

Multimodal Large Language Models (MLLMs) exhibit impressive capabilities in multimodal tasks such as
image understanding,
visual question answering, and 
instruction following~\citep{liu2024llava, gpt4v, tong2024cambrian}.
These models leverage the strong reasoning abilities of LLMs~\citep{openai2023gpt, zheng2024judging,llama3}, with
advancements in MLLMs driven primarily by scaling up the language models~\citep{liu2024llava, scalingllm}.
The rapid progress in model capabilities necessitates the development of more stronger benchmarks that are however missing.
Recent works such as Cambrian-1~\citep{tong2024cambrian}, OpenEQA~\citep{openeqa}, and MMStar~\citep{blind_mllm_eval} reveal that current benchmarks~\citep{hiippala2021ai2d,lu2023mathvista, yue2024mmmu} exhibit a strong language bias and fail to accurately assess the visual understanding capabilities of MLLMs.
This motivates us to further explore the vision-centric evaluation of these models.

Since MLLMs are conversational agents, users can evaluate the visual understanding of these models through Visual Question Answering (VQA), by examining their natural language responses.
However, reliably evaluating natural language responses requires commonsense reasoning and language comprehension.
Although LLMs can be used for this purpose, they may be inaccurate and slow when parsing a large number of responses (\eg~LLaMA-3-8B takes \SI{3.1}{\second} to generate 100 tokens on a single A6000 GPU).
To circumvent this, recent works~\citep{tong2024cambrian, mmvp, openeqa, fu2024blink} frame fine-grained visual tasks as VQA, where the model is required to \textit{select an answer} from the multiple options.

\begin{figure}[t]
\centering
\includegraphics[height=0.88\textheight]{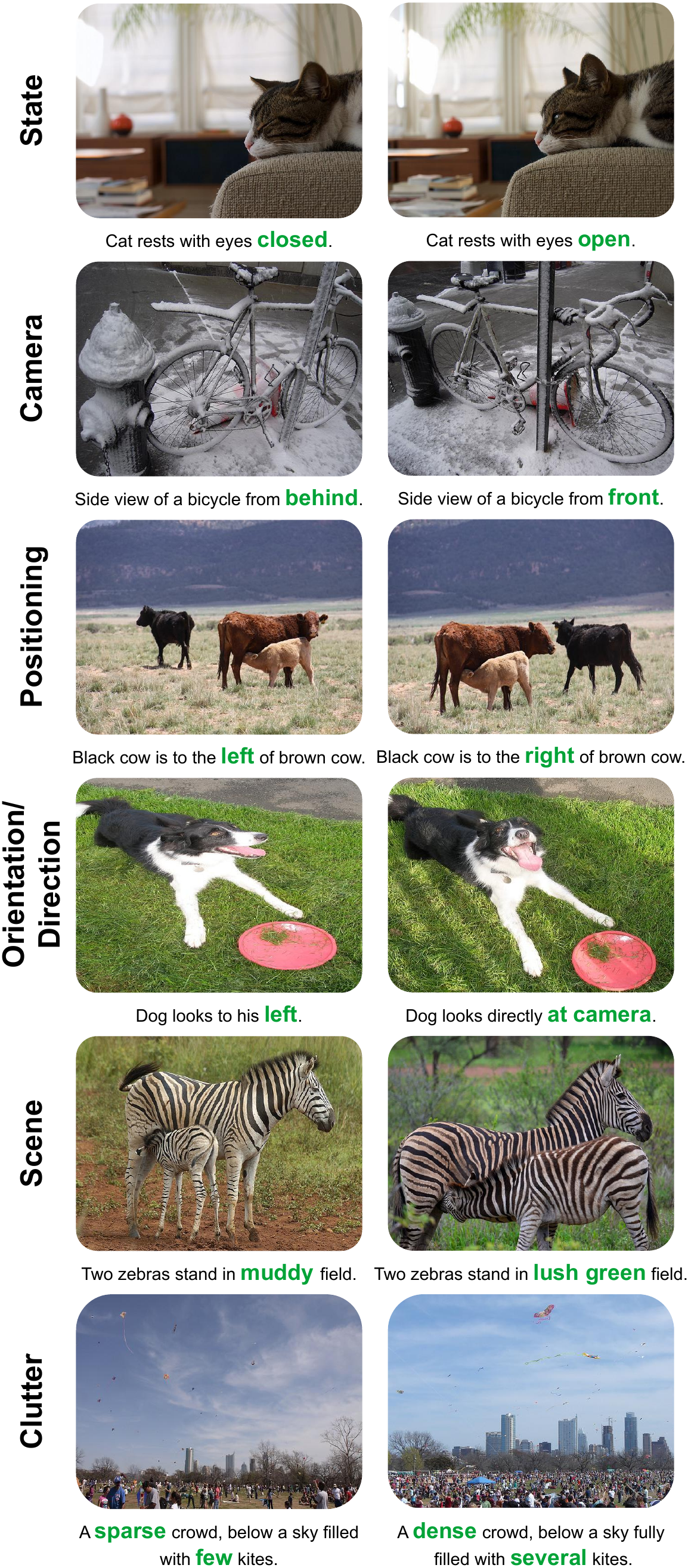}
\caption{
Each row illustrates one of the six Points of Difference (PODs) present in \benchmark{} benchmark:
(a)~State,
(b)~Camera,
(c)~Positioning,
(d)~Orientation/Direction,
(e)~Scene, and
(f)~Clutter.
We provide exemplar discriminative captions for each image, highlighting the fine-grained point of difference in \textbf{\textcolor{ForestGreen}{green}}.}
\vspace{-8mm}
\label{fig:teaser_pod}
\end{figure}

VQA with multiple choices, provides a reliable method for checking the existence of specific facets of visual understanding.
For instance, we can assess the MLLM's ability to capture the \textit{state} of the object in Fig.~\ref{fig:teaser_pod}a by directly asking it if the cat's eyes are \textit{open} or \textit{closed}.
However, we find that it is \textit{easier for the model to select an answer from multiple choices than to generate the answer} itself (see Fig.~\ref{fig:vqa_teaser}).
Specifically, providing the answer along with task prompt (through multiple choice options or as part of the question) biases the MLLM's output towards the visual concept that is being evaluated.
This raises questions about whether the model actually understands this visual concept or is simply picking the more likely choice.

In this work, we offer an alternative perspective for evaluating fine-grained understanding exhibited by an MLLM.
While looking at a pair of images, we ask the model to describe the target image such that a listener can distinguish the target image from an extremely similar distractor~\citep{gaur2024, concadia}. 
We curate extremely similar image pairs, each having one prominent point of visual difference such that solving this task entails that the model captures a specific facet of visual understanding.
With 247 such image pairs, we introduce the \textbf{\benchmark} benchmark (examples in Fig.~\ref{fig:teaser_pod}).
For each image pair, we prompt the model to:
(1)~\textbf{Detect} the visual difference, and
(2)~\textbf{Describe} the target image uniquely such that it
(3)~\textbf{Discriminate}s the target image from the distractor.

Unlike VQA, we do not constrain the output to multiple-choice answers and directly evaluate the model's free-form natural language generation.
Specifically, we assess the ability of the MLLM to capture subtle visual distinctions using self-retrieval~\citep{gaur2024, liu2018selfretrieval}, \ie retrieving the target image based on its generated description against the distractor image.
Although \citet{gaur2024} also use self-retrieval for fine-grained evaluation, the captioner does not have access to  the distractor images.
In contrast, we evaluate the MLLM's ability to discern fine-grained visual differences by showing both images simultaneously.

Inspired by MMVP~\citep{mmvp}, we annotate each image pair with a specific point of difference (POD), highlighting the visual concept that distinguishes both images.
\benchmark{} consists of image pairs that can be distinguished primarily using one of the following PODs: state, position, scene, orientation/direction, camera, or clutter (Sec.~\ref{subsec:annotating_pod}).
Consequently, self-retrieval can be used as a white-box evaluation to assess different facets of visual understanding captured by the MLLM (Sec.~\ref{subsec:wbe}).
For example, successful self-retrieval for the image pair in Fig.~\ref{fig:teaser_pod}d requires the model to have a fine-grained understanding of the dog's \textit{orientation}.

Although CounterCurate~\citep{zhang2024countercurate} and Spot-the-Diff~\citep{jhamtani2018learning} also elicit fine-grained visual discrimination with image pairs, CounterCurate~\citep{zhang2024countercurate} uses VQA with synthetic images, while Spot-the-Diff~\citep{jhamtani2018learning} leverages nearby frames from video-surveillance footage that often lack clear semantic differences beyond \textit{negation} and \textit{object position}.
Most similar to us, MMVP~\citep{mmvp} evaluates the ability of MLLMs to discern fine-grained visual distinctions across image pairs through VQA.
While both tasks require the model to identify a specific visual difference, in the case of MMVP, the visual concept under scrutiny is presented to the MLLM within multiple choice options or as part of the question itself (see Fig.~\ref{fig:vqa_teaser}).
In contrast, \benchmark{} requires the model to \textit{independently detect} the visual difference and incorporate it into captions that uniquely describe each image, making it a more challenging task than VQA (see \Cref{app:vqa_easier}).

Unlike VQA, which offers flexibility through the phrasing of questions, self-retrieval provides flexibility through manual curation of image pairs.
Specifically, one can assess a model's ability to discern specific aspects of visual discrimination by curating image pairs that differ only in the targeted visual concept.
As a result, image pairs in \benchmark{} are significantly more challenging to distinguish.
We verify this by using Gemini-1.5-Pro to perform self-retrieval evaluation on image pairs from both datasets under the same experimental settings --
the model scores 87.3\% in MMVP, compared to a mere 39.7\% on our \benchmark{} benchmark.

We evaluate various open- and closed-source MLLMs on our benchmark (Sec.~\ref{subsec:results}).
We find that current models struggle in capturing fine-grained visual differences, with open-source models fail to outperform random guess.

\section{\benchmark{} Benchmark}
\label{sec:benchmark}


We present how we curate image pairs in the \benchmark{} benchmark followed by an explanation of each point of difference.

\subsection{Curating Image Pairs}
Our benchmark creation process requires densely captioned images.
We source them in two ways:
(1)~ShareGPT4V~\cite{sharegpt4v} that comprises 100K images with dense captions generated using GPT-4V~\cite{gpt4v}, and
(2)~HolisticCaps~\cite{gaur2024} that blends multiple human annotated COCO captions and combines them with dense visual descriptions from InstructBLIP~\cite{instructblip}.
Lengthy captions are summarized to 65 tokens using Llama-3-70B~\cite{llama3} so they do not exceed SigLIP~\cite{siglip} text encoder's token capacity.
We briefly summarize the image pair curation process (please refer to~\citep{gaur2024} for a more detailed description):
(1)~The images and their corresponding (summarized) captions are encoded and concatenated using SigLIP's image and text encoder respectively.
(2)~Image pairs are identified based on highest similarity in the multimodal embedding space ensuring that each pair encodes a unique visual concept.
(3)~Finally, we perform manual filtering
to get 247 image pairs, with each image pair having one prominent point of difference.

\subsection{Annotating Points of Difference (POD)}
\label{subsec:annotating_pod}

Each image pair is essentially identical except for one prominent visual difference.
We annotate the differences among 6 visual concepts.
The number of image pairs differentiated by each concept is mentioned in parentheses.

\begin{itemize}
\setlength\itemsep{2mm}
\item \textbf{State} (72):
relates to different states of the same object, \eg the toilet seat is \textit{up/down}. This also includes entities performing different actions, \eg, in Fig.~\ref{fig:teaser_pod}a, we see a cat \textit{rests} with eyes \textit{open/closed}.
\item \textbf{Camera} (55):
relates to the camera's position or different properties such as \textit{perspective}, \textit{viewpoint}, \textit{depth}, or \textit{zoom}.
Fig.~\ref{fig:teaser_pod}b shows an example of a different viewpoint.
\item \textbf{Position} (26):
relates to different relative positions of objects in an image.
\Eg the black cow is on the \textit{left/right} of the brown cow in Fig.~\ref{fig:teaser_pod}c.
This also includes cases where a \textit{single} object has different relative positioning with respect to the background.
\item \textbf{Orientation/Direction} (63):
relates to the entity or object facing a different direction or having dissimilar orientation with respect to the camera.
Fig.~\ref{fig:teaser_pod}d shows an example of a dog looking left or at the camera.
\item \textbf{Scene} (18):
relates to fine-grained difference in attributes or background characteristics that span the image or some area around an object of interest.
\Eg, Fig.~\ref{fig:teaser_pod}e shows two zebras standing in a \textit{lush green/muddy} environment. 
\item \textbf{Clutter} (13):
relates to a pair of images with extremely similar scenes, \eg a fridge or a desk, cluttered with too many objects. 
The image pair often contains fine-grained differences in the characteristics of the non-prominent objects or scene.
An example of this is the \textit{dense/sparse} crowd of people in Fig.~\ref{fig:teaser_pod}f. 
\end{itemize}

\paragraph{Inter-annotator agreement}
is studied by randomly sampling 100 image pairs from the benchmark and asking two annotators to pick the most relevant POD based on the original instructions.
After removing some invalid annotations, we obtain an agreement accuracy of 71.3\% (67/94), indicating that the PODs are reasonably distinct.


\section{Experiments}
\label{sec:experiments}

\subsection{Self-Retrieval Setup}

Given an image pair $I_0$ and $I_1$, we prompt an MLLM to generate image captions $C_0$ and $C_1$ that allow a listener to distinguish between them.
To evaluate the MLLM, we first encode both the images and the generated captions using \texttt{siglip-so400m-patch14-384}.
Next, similar to Winoground~\cite{winoground}, we compute the \textit{self-retrieval score} that checks whether the scorer is able to pair the generated caption to the correct target image:
\begin{equation}
f(C_0, I_0, C_1, I_1) = 
\begin{cases}
1 & \text{if } sim(C_0, I_0) > sim(C_0, I_1) \\
& \text{and } sim(C_1, I_1) > sim(C_1, I_0), \\
0 & \text{otherwise,}
\end{cases}
\end{equation}
where $sim(C_i, I_j)$ corresponds to the cosine similarity between the encoded representations of the caption and image respectively.

\subsection{Prompting MLLMs}
Both images are simultaneously given to the MLLM with the prompt shown in \cref{fig:prompt}.
Notably, since Cambrian~\cite{tong2024cambrian} and LLaVA-NeXT~\cite{liu2024llava} do not support multi-image prompting, we concatenate the image pair horizontally (without resizing). 
Extensive prompt tuning is done for open-source models such as Cambrian-34B, Chameleon-30B~\cite{team2024chameleon}, and LLaVA-NeXT-34B to improve performance on the self-retrieval task (see \Cref{app:prompts}).

\begin{figure}[h]
\centering
\noindent\begin{minipage}{0.99\linewidth}
\mdfsetup{%
middlelinewidth=1pt,
backgroundcolor=green!3,
innerleftmargin=0.2cm,
innerrightmargin=0.2cm,
roundcorner=12pt}
\begin{mdframed}
\vspace{0.2em}
\texttt{\footnotesize\linespread{0.2}\textbf{Prompt:} Identify fine-grained visual differences between both images and generate a discriminant caption for each image.
Each caption should uniquely describe the image and highlight the distinct features that set the image apart from the other.
Output in JSON format with `left' and `right' as keys, and their captions as string values.}
\end{mdframed}
\end{minipage}
\caption{The prompt given to GPT 4o, Claude Sonnet 3.5, Gemini Pro 1.5, and Gemini Flash 1.5 to uniquely describe images within our benchmark.
Prompts for the open-source models are presented in \Cref{app:prompts}.
}
\label{fig:prompt}
\end{figure}

\begin{figure}[t]
\centering
\includegraphics[width=1\linewidth]{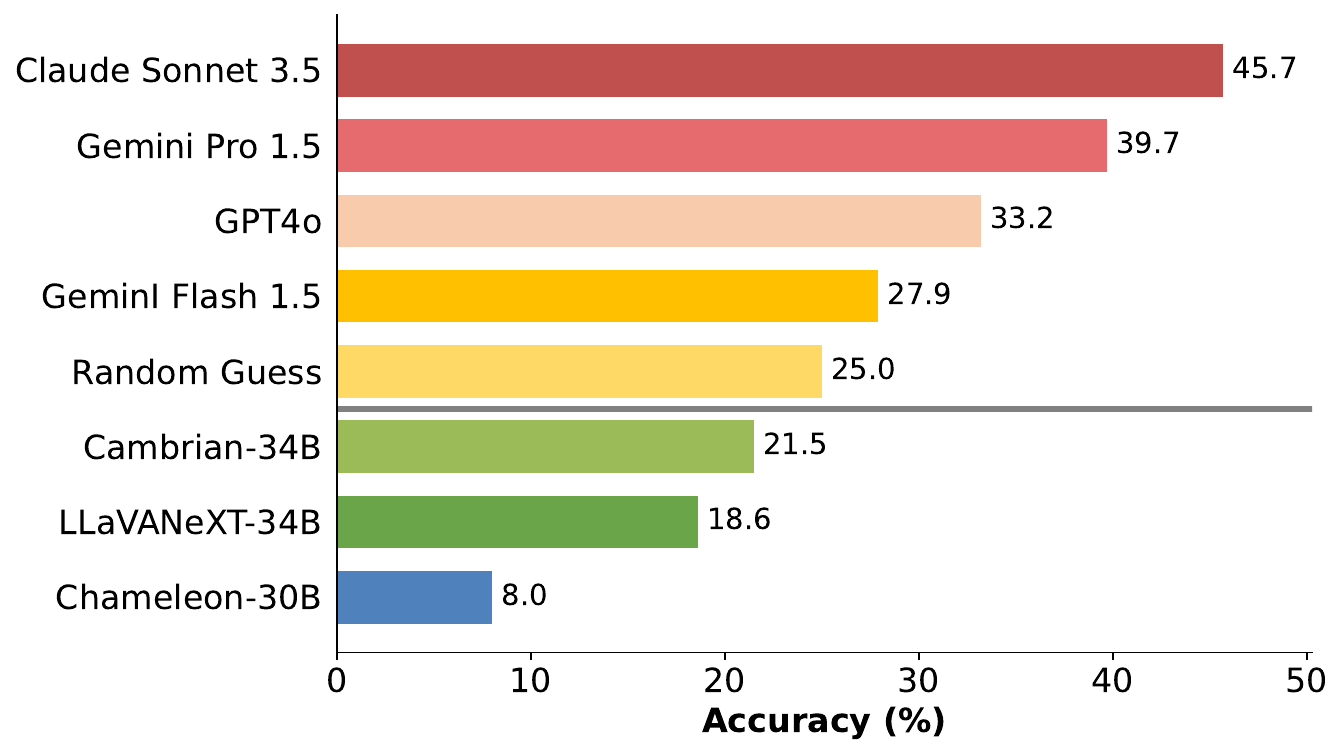}
\vspace{-5mm}
\caption{Benchmarking different open- and closed-source MLLMs on \benchmark{} benchmark. The scores are averaged across all image pairs.}
\label{fig:baselines}
\vspace{-5mm}
\end{figure}

\begin{figure*}[t]
\centering
\includegraphics[width=0.8\textwidth]{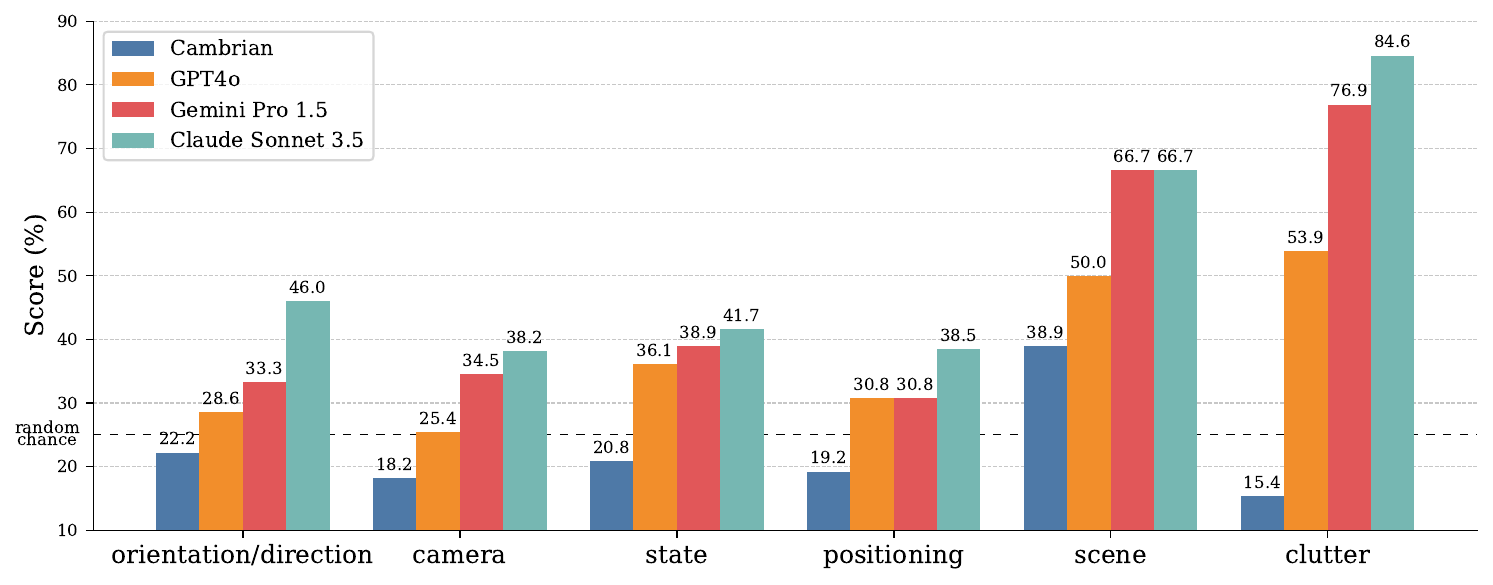}
\vspace{-2mm}
\caption{\textbf{Whitebox evaluation on \benchmark}. We compute self-retrieval scores on individual \textit{POD} subsets within our benchmark for Cambrian-34B, GPT 4o, Gemini Pro 1.5, and Claude Sonnet 3.5. The random guess is 25\%.
}
\vspace{-2mm}
\label{fig:whitebox_eval}
\end{figure*}

\subsection{MLLMs struggle with fine-grained Differences}
\label{subsec:results}
The performance of MLLMs on \benchmark{} benchmark is shown in Fig.~\ref{fig:baselines}.
MLLMs struggle to incorporate fine-grained visual details in their captions, as evidenced by their performance on the benchmark. 
State-of-the-Art (SotA) open-source models such as Cambrian-34B and LLaVA-NeXT-34B fail to even outperform random guess. 
Although, closed-source models outscore random guess, they still struggle to generate discriminant captions for images in \benchmark, with GPT-4o achieving only 33.2\%.
We find Claude Sonnet 3.5 to be most capable in discerning fine-grained visual differences, achieving the highest score of 45.7\% on our benchmark.

\subsection{Whitebox Evaluation}
\label{subsec:wbe}
Each image pair within \benchmark{} contains a prominent visual concept or point of difference that.
Generating discriminative descriptions for both images requires the caption to incorporate the targeted visual difference.
When an MLLM fails to uniquely describe and retrieve both images within a pair in \benchmark, we are able to accurately identify the visual concept that the MLLM is unable to pick up (see \Cref{app:quali_pod} for qualitative examples).

Fig~\ref{fig:whitebox_eval} illustrates the self-retrieval performance of various MLLMs on image pairs of \benchmark{} across all six points of difference. 
Trends indicate that SOTA MLLMs, both open and closed, struggle to perceive fine-grained changes in orientation/direction, camera angle, object's state, or positioning.
These findings are similar to those established in MMVP~\cite{mmvp}.
In contrast, uniquely describing similar images with differing scenes appears to be easier for these models. 
This may be because differences in the scene often span the entire image which is easier for current models to identify.
V\textsuperscript{*}~\cite{wu2024vstar} finds that MLLMs struggle to focus on fine-grained details in visually crowded images. 
Interestingly, while we find this to be true for Cambrian-34B, Fig~\ref{fig:whitebox_eval} demonstrates that closed-source MLLMs, specifically Claude 3.5 Sonnet and Gemini Pro 1.5, are capable of identifying characteristics of non-prominent objects to distinguish cluttered images.

\subsection{Validity of Self-Retrieval Scorer}
While self-retrieval is able to assess an MLLM's ability to detect and describe subtle visual distinctions between images, quantifying its success depends on the ability of the scorer function used for retrieval.
Even if an MLLM successfully captures and incorporates visual differences in its captions, an inferior scorer may fail to capture these fine-grained details, leading to unreliable pairings between generated captions and images.
To address this challenge as effectively as possible, we adopt SigLIP, one of the most fine-grained open-source VLM, as our scorer.

We conduct a study to evaluate the reliability of SigLIP as a scorer for the self-retrieval task by comparing three scorers: an average human, an expert human, and SigLIP.
Each scorer is presented 100 image pairs sampled from the benchmark. 
We use captions generated by GPT-4o and pick one (among two) randomly as the caption of the target image.
From the 100 captions generated using GPT-4o, 23 captions are deemed non-discriminant by the expert human scorer and are filtered out.

Among the remaining 77 captions, a text-to-image retrieval task is set up: within each image pair, all three scorers are asked to retrieve the target image using the given caption.
We find an agreement of 94.8\% between the average human and expert human scorer, with the average human picking the same image as the expert for 73/77 samples.
SigLIP shows a 79.2\% agreement with the expert human scorer, matching the image selection for 61/77 image pairs.
This suggests that although SigLIP is not a perfect scorer, it is good enough to be used for self-retrieval evaluation on our benchmark.

\section{Conclusion}
\label{sec:conclusion}

Our study sheds light on the capabilities as well as limitations of current Multimodal LLMs (MLLMs) in perceiving and describing subtle visual differences.
We propose a novel benchmark, \benchmark, consisting of 247 image pairs, each comprised of highly similar images that differ in one fine-grained visual concept. 
By using self-retrieval, we directly evaluate the natural language outputs requiring the model to independently identify the visual difference and incorporate it into unique captions.
Our study reveals that while MLLMs excel at distinguishing scene changes and visually crowded images, they struggle with nuanced aspects of visual understanding such as camera angle or an object's state, positioning, and orientation. 
Among all the MLLMs tested, Claude Sonnet 3.5 performs relatively well on \benchmark, while other models, especially open-source ones, struggle.
We also conduct a human study to investigate the validity of the retrieval scorer. 
Our results reveal SigLIP to be a reliable scorer for self-retrieval evaluation on \benchmark.

Future work could focus on scaling up the benchmark by adopting large image datasets with accompanying dense captions such as PixelProse~\cite{pixelprose}, VeCap~\cite{veclip}.
This could result in a larger number of fine-grained image pairs, enhancing diversity and reliability of evaluation.

\paragraph{Acknowledgements.}
This project was supported in part by funding from SERB SRG/2023/002544 and a research gift by Adobe Research India.

{
\small
\bibliographystyle{ieeenat_fullname}
\bibliography{bib/longstrings,bib/main}
}
\clearpage
\appendix
\onecolumn
\section{Appendix}
\label{sec:appendix}
\subsection{Prompt Tuning for Open-source MLLMs}
\label{app:prompts}

Open-source MLLMs required further prompt tuning compared to closed-source MLLMs. 
The following prompt was used for benchmarking Cambrian-34B and LLaVA-NeXT-34B on \benchmark:

\noindent
\texttt{\small \linespread{0.1}
Given two images, generate a discriminant description for each image that uniquely describes it. The description for each image should highlight its key visual differences compared to the other distractor image. Only focus on distinct visual features that make each image unique.
Output in JSON format, with a key `left' and a key `right' with string values corresponding to descriptions for the left and right image.} \\

Chameleon-30B struggled to follow the above instructions, and its prompt was tuned further. 
We found the following prompt to perform best on our benchmark:

\noindent
\texttt{\small \linespread{0.1}
You are given two images. Identify finegrained visual differences between both images <image><image>. Using these visual differences, uniquely describe each image. The output should be in JSON format, with 'image1' and 'image2' as keys, and their respective captions as string values.}


\subsection{VQA is easier than \benchmark{} evaluation}
\label{app:vqa_easier}

\begin{figure}[b]
\centering
\includegraphics[width=0.8\textwidth]{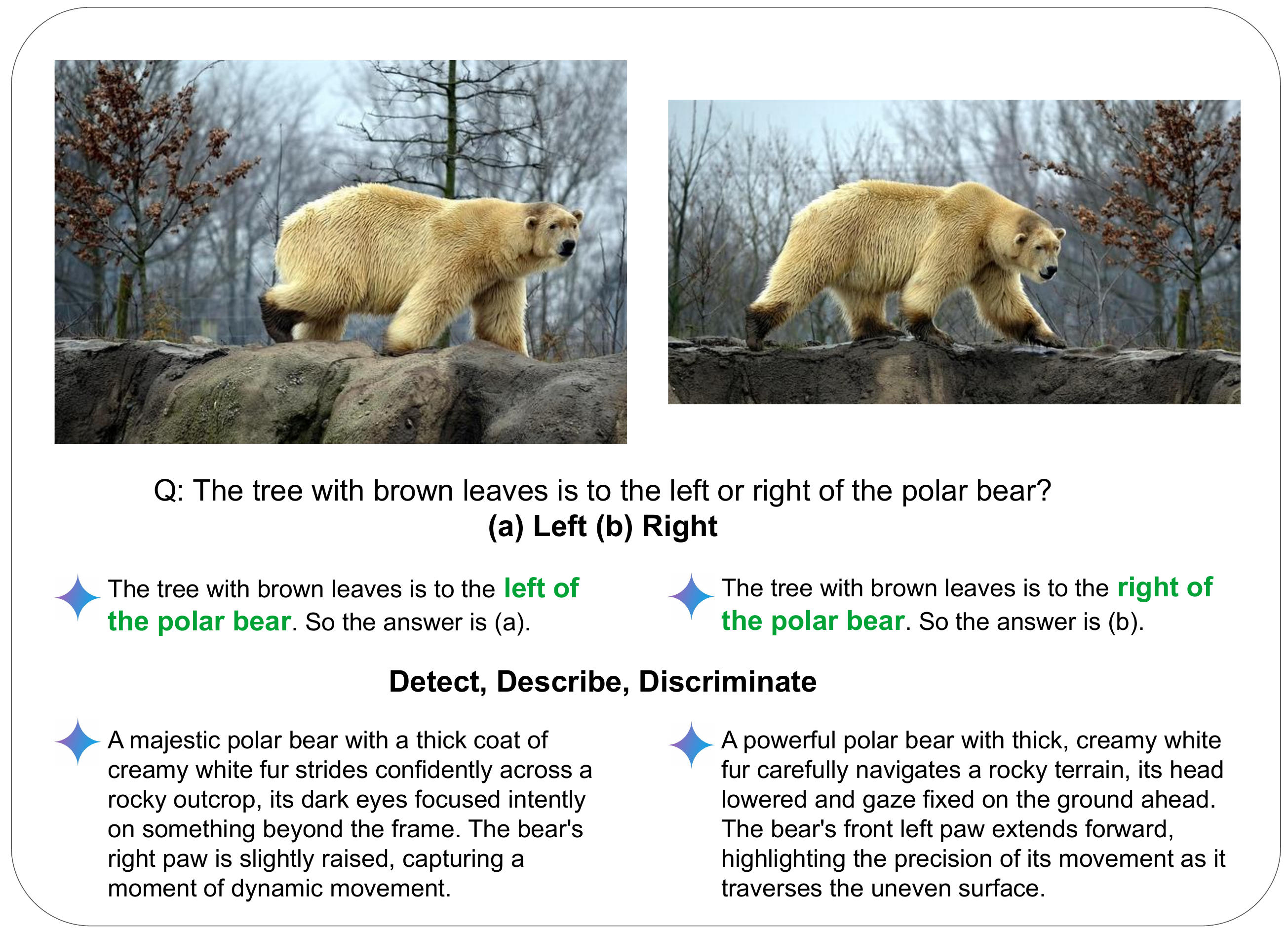}
\vspace{-2mm}
\caption{\textbf{VQA \vs~\benchmark{} evaluation, 1/4.}
Example from \benchmark{}, where model fails during \benchmark{} evaluation despite achieving perfect score in VQA.}
\label{fig:vqa_d3_1}
\vspace{-2mm}
\end{figure}

\begin{figure}[t]
\centering
\includegraphics[width=0.8\textwidth]{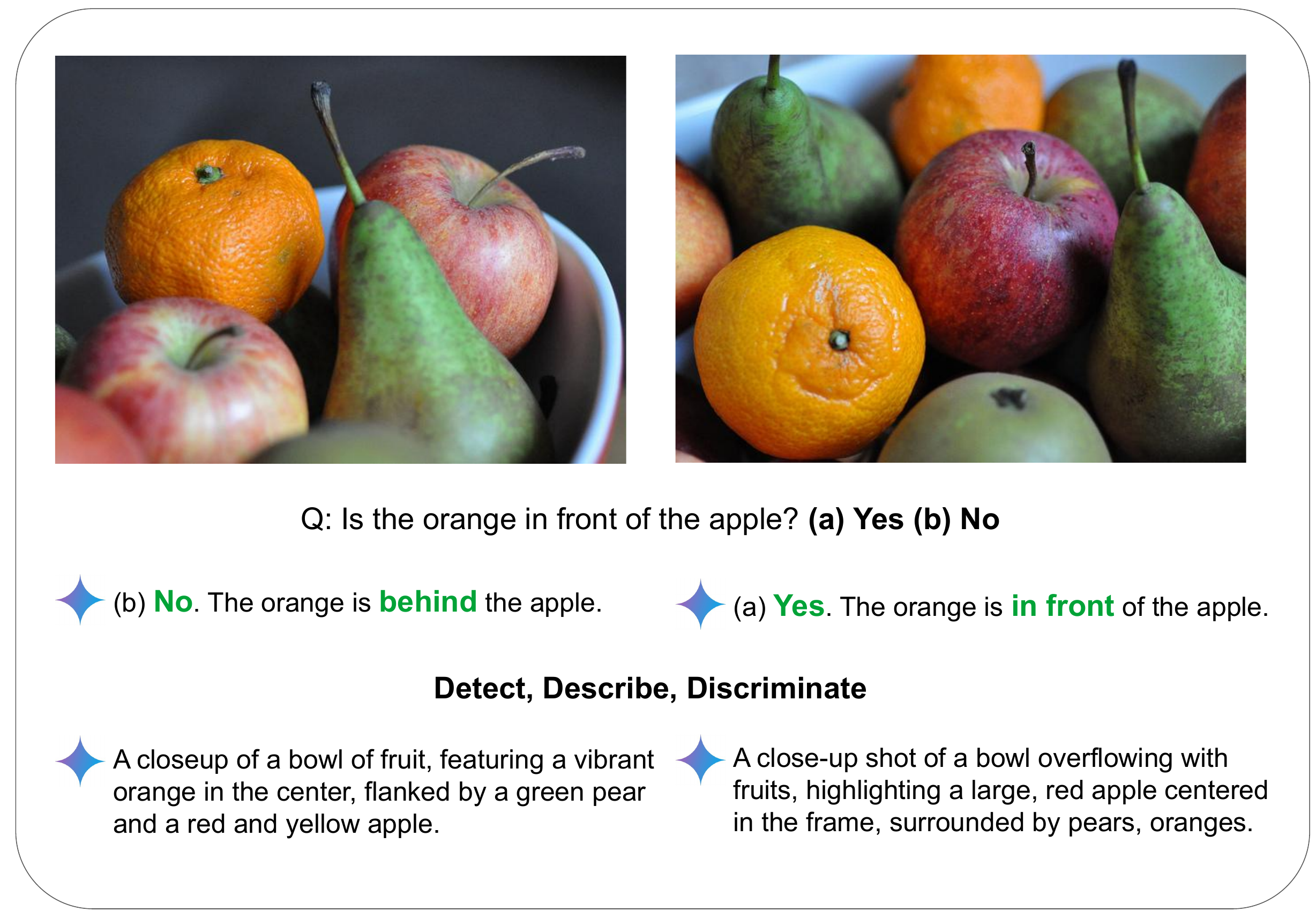}
\vspace{-2mm}
\caption{\textbf{VQA \vs~\benchmark{} evaluation, 2/4.}
Example from \benchmark{}, where model fails during \benchmark{} evaluation despite achieving perfect score in VQA.}
\label{fig:vqa_d3_2}
\vspace{-2mm}
\end{figure}

\begin{figure}[t]
\centering
\includegraphics[width=0.8\textwidth]{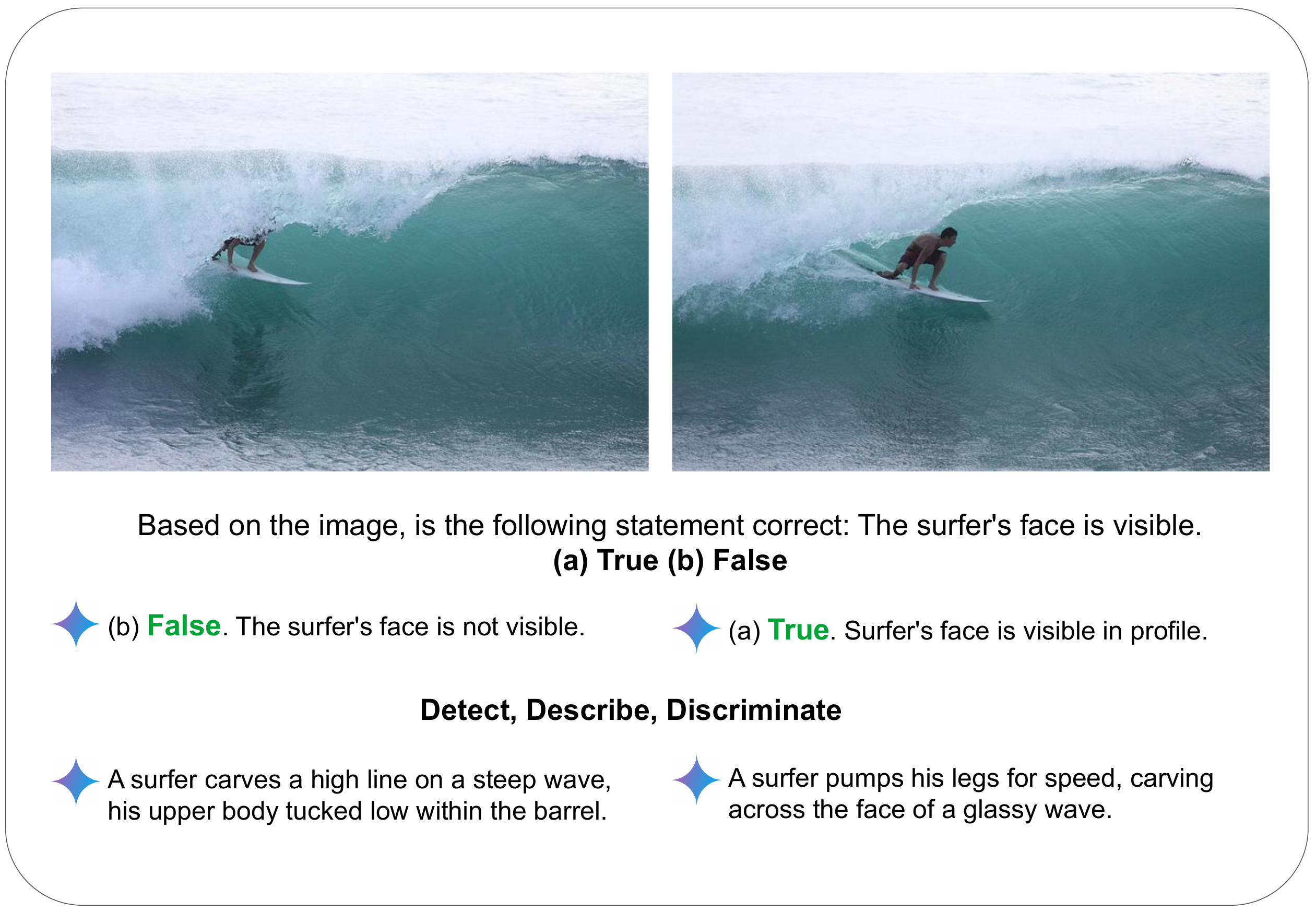}
\vspace{-2mm}
\caption{\textbf{VQA \vs~\benchmark{} evaluation, 3/4.}
Example from \benchmark{}, where model fails during \benchmark{} evaluation despite achieving perfect score in VQA.}
\label{fig:vqa_d3_3}
\vspace{-2mm}
\end{figure}

\begin{figure}[t]
\centering
\includegraphics[width=0.8\textwidth]{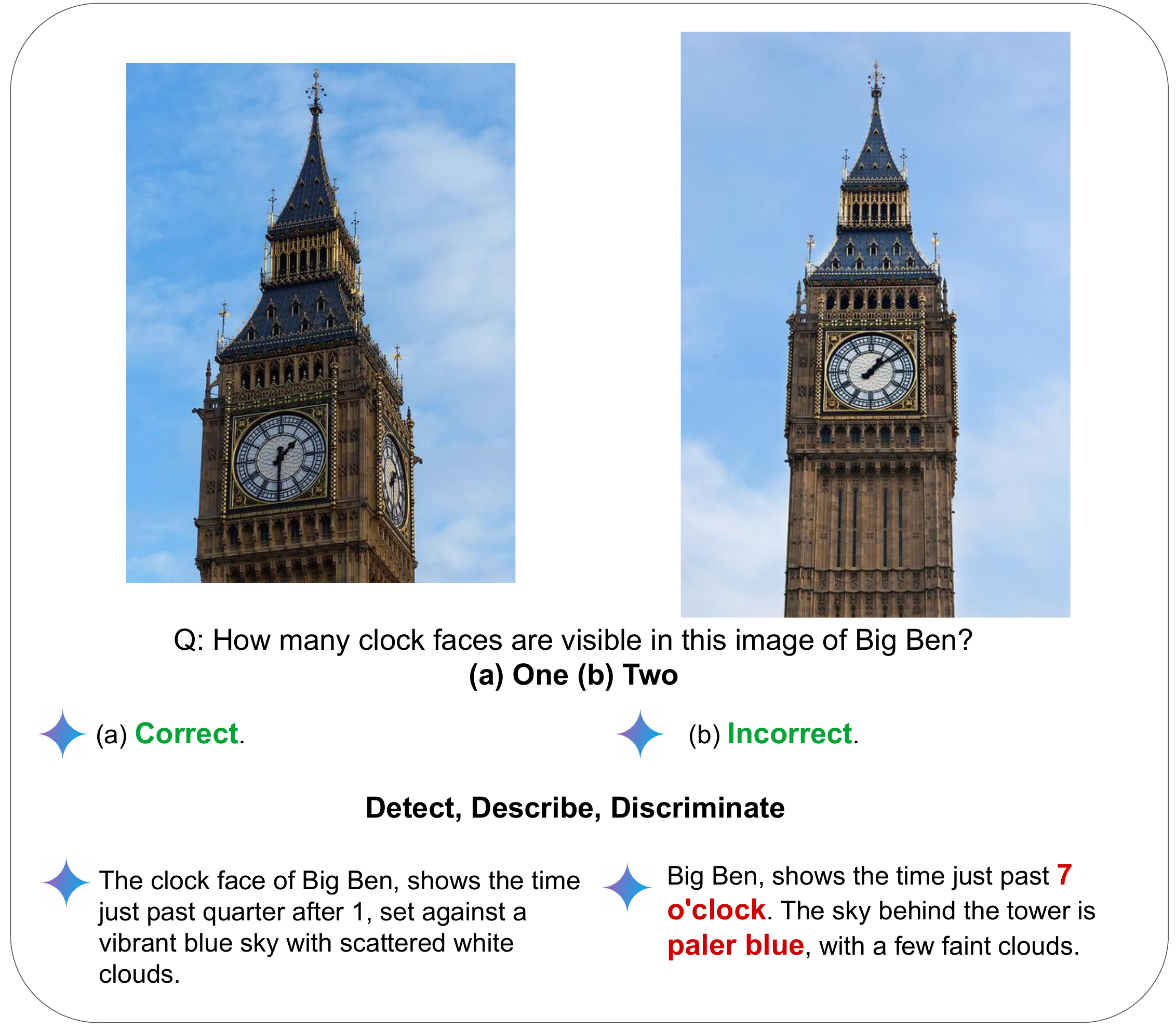}
\vspace{-2mm}
\caption{\textbf{VQA \vs~\benchmark{} evaluation, 4/4.}
Example from \benchmark{}, where model fails during \benchmark{} evaluation despite achieving perfect score in VQA.}
\label{fig:vqa_d3_4}
\vspace{-2mm}
\end{figure}

Given an image pair, we find that it is easier for the MLLM to discern fine-grained differences during VQA than \benchmark{} evaluation.
Although providing the visual concept as part of the question aids VQA in assessing the existence of specific visual abilities, it makes the task easier.
Figs.~\ref{fig:vqa_d3_1}, \ref{fig:vqa_d3_2}, \ref{fig:vqa_d3_3}, \ref{fig:vqa_d3_4} provide qualitative examples where Gemini-1.5-Pro, despite achieving perfect VQA scores, struggles to independently identify the fine-grained difference and incorporate them into discriminant captions.
In the figures, the text highlighted in \textbf{\textcolor{ForestGreen}{green}} represents the fine-grained differences´ captured by the MLLMs, while that marked in \textbf{\textcolor{BrickRed}{red}} represents erroneous descriptions (hallucinations).

\subsection{Qualitative Analysis of MLLMs on Different Points of Difference}
\label{app:quali_pod}

\begin{figure}[t]
\centering
\includegraphics[width=0.8\textwidth]{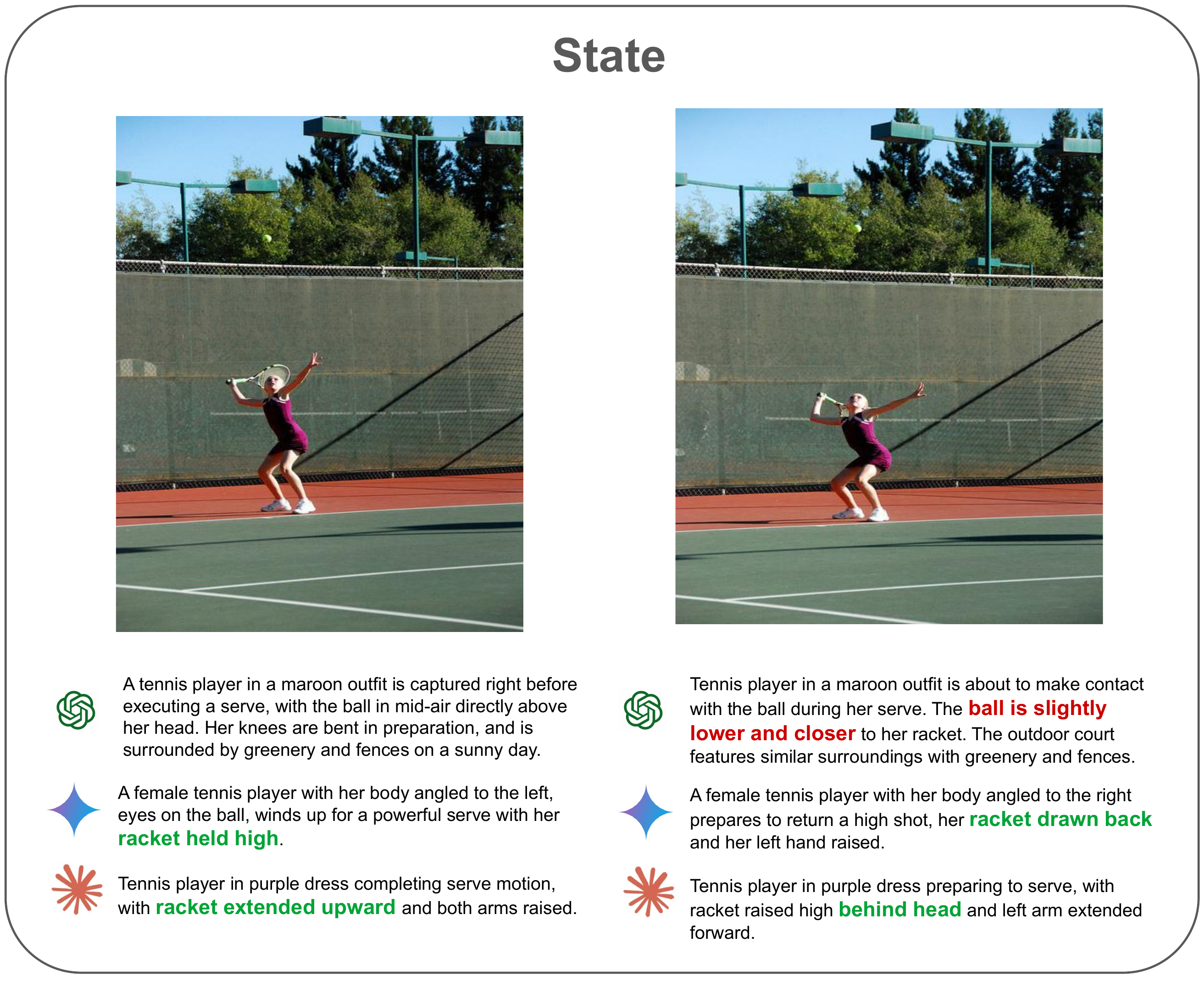}
\vspace{-2mm}
\caption{Outputs of different closed-source MLLMs during \benchmark{} evaluation on an image pair having \textbf{State} as POD.}
\vspace{-2mm}
\label{fig:whitebox_eval_state}
\end{figure}

\begin{figure}[t]
\centering
\includegraphics[width=0.8\textwidth]{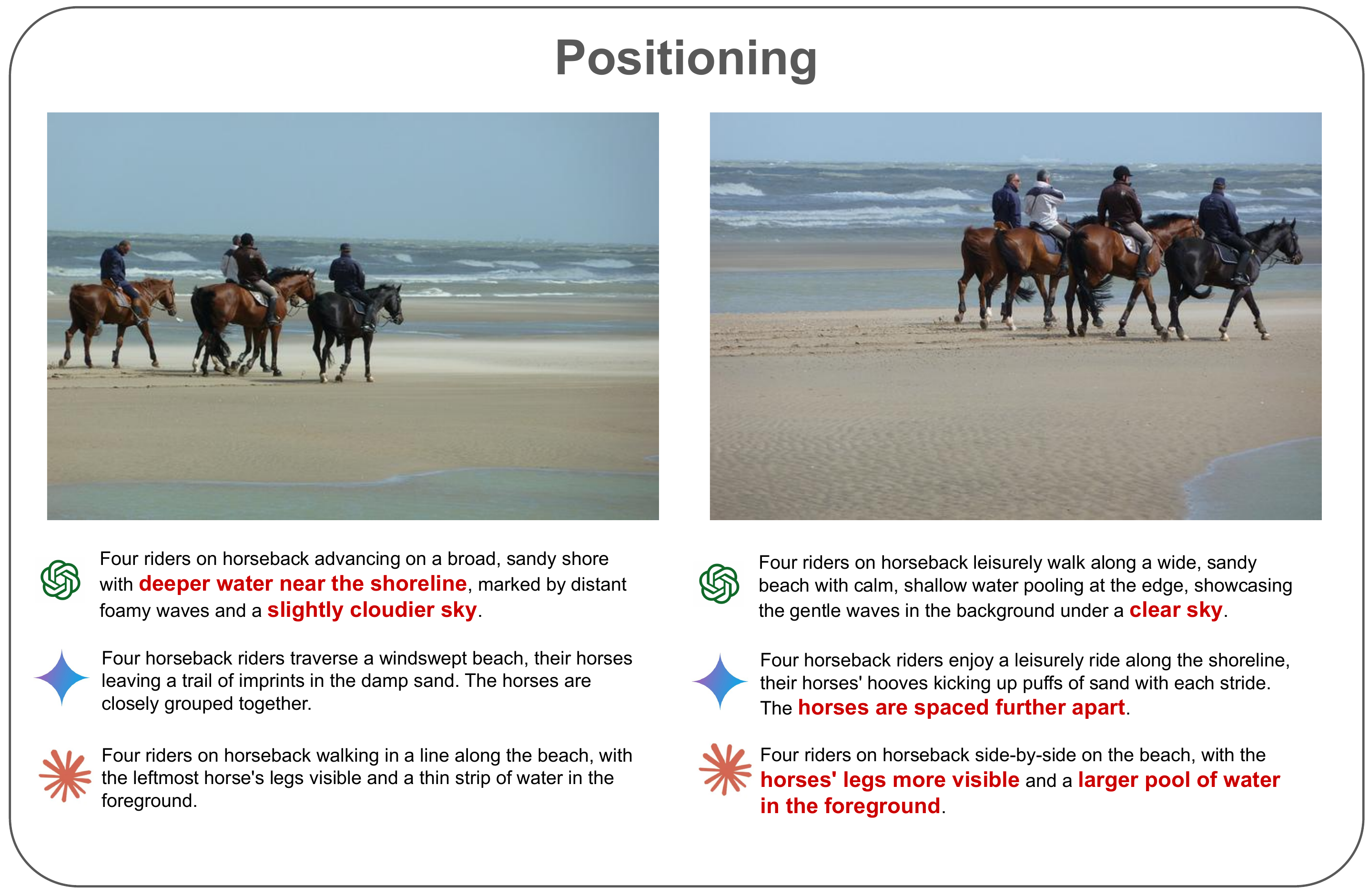}
\vspace{-2mm}
\caption{Outputs of different closed-source MLLMs during \benchmark{} evaluation on an image pair having \textbf{Positioning} as POD.}
\vspace{-2mm}
\label{fig:whitebox_eval_positioning}
\end{figure}

\begin{figure}[t]
\centering
\includegraphics[width=0.8\textwidth]{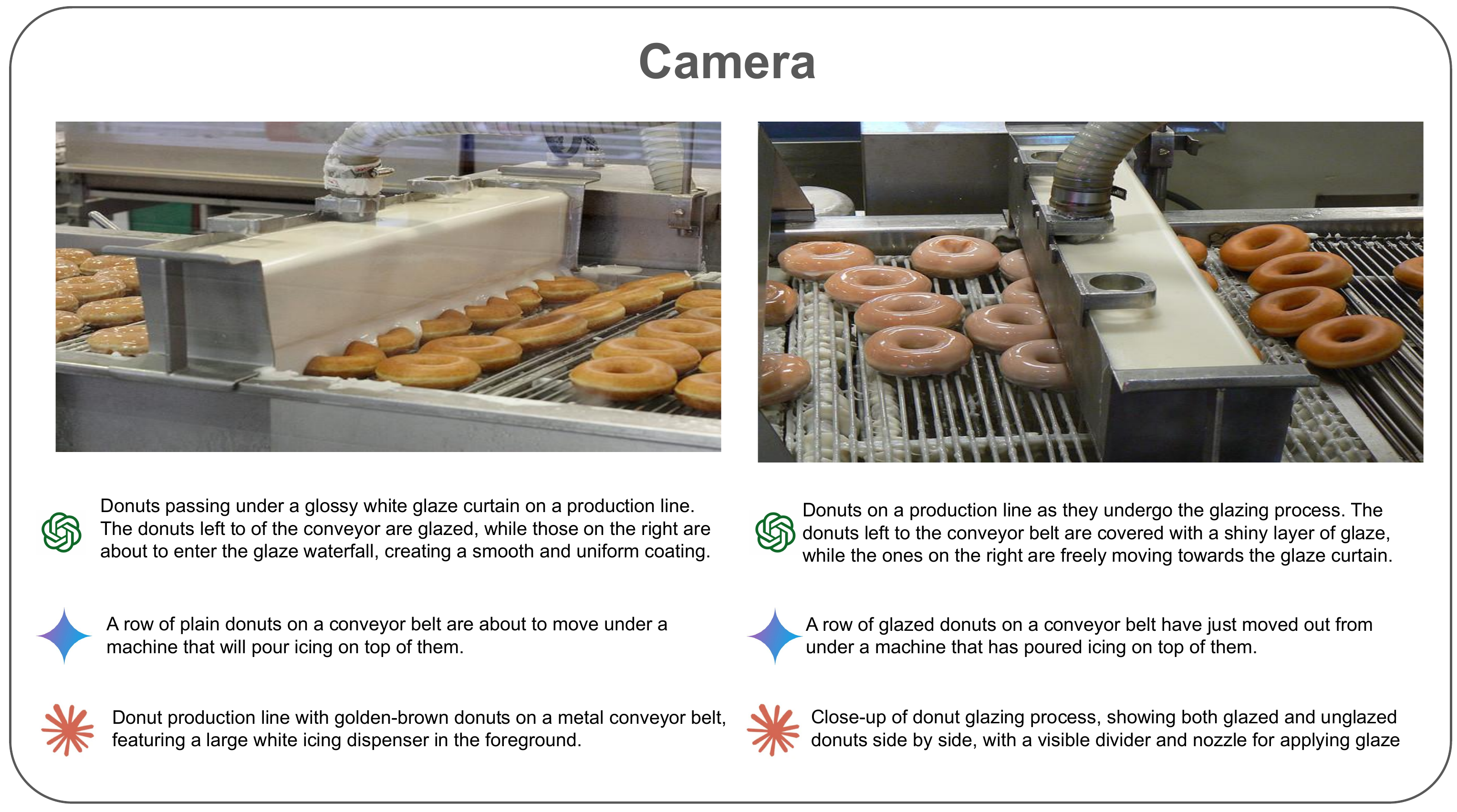}
\vspace{-2mm}
\caption{Outputs of different closed-source MLLMs during \benchmark{} evaluation on an image pair having \textbf{Camera} as POD.}
\vspace{-2mm}
\label{fig:whitebox_eval_camera}
\end{figure}

\begin{figure}[t]
\centering
\includegraphics[width=0.8\textwidth]{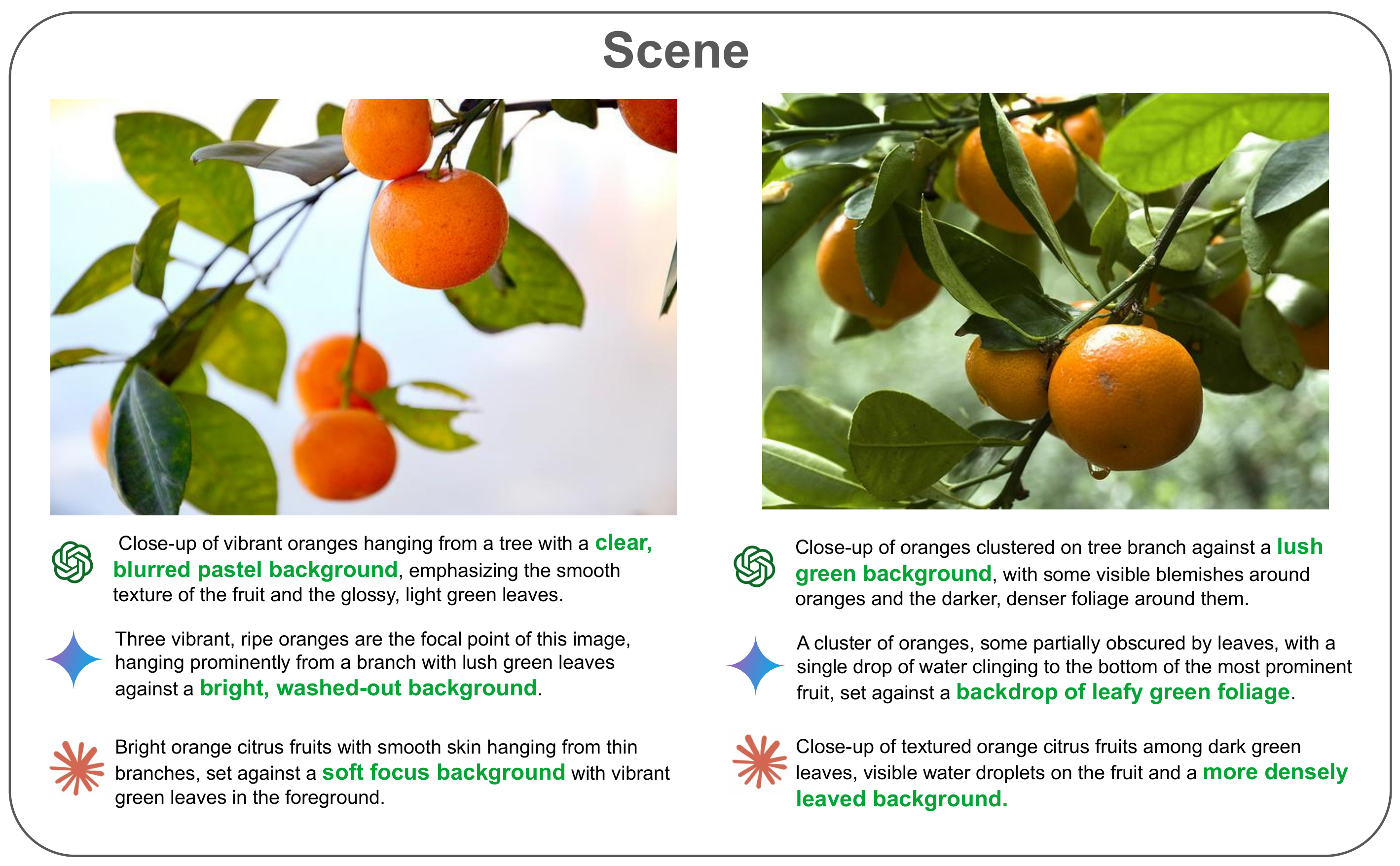}
\vspace{-2mm}
\caption{Outputs of different closed-source MLLMs during \benchmark{} evaluation on an image pair having \textbf{Scene} as POD.}
\vspace{-2mm}
\label{fig:whitebox_eval_scene}
\end{figure}

\begin{figure}[t]
\centering
\includegraphics[width=0.8\textwidth]{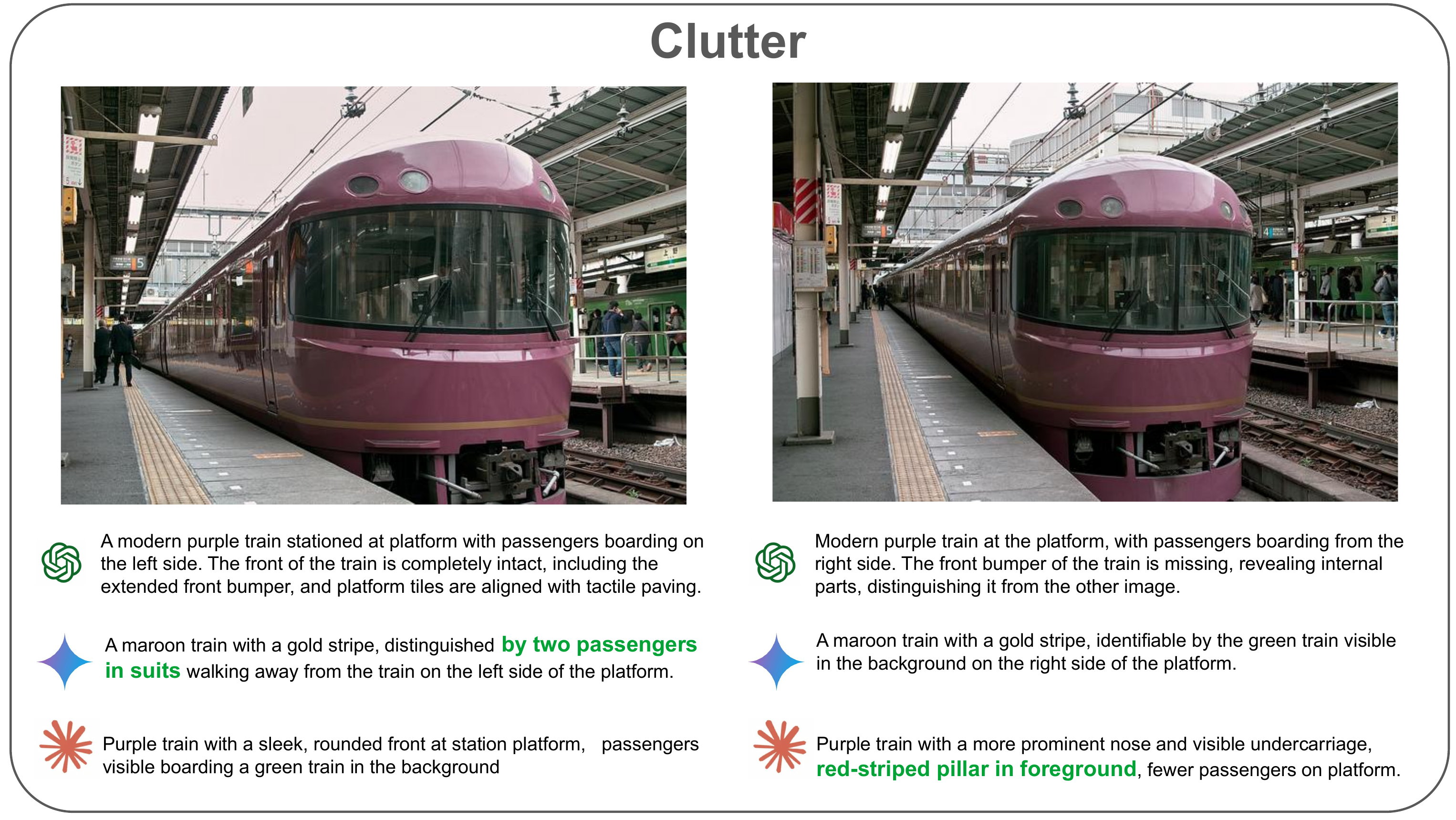}
\vspace{-2mm}
\caption{Outputs of different closed-source MLLMs during \benchmark{} evaluation on an image pair having \textbf{Clutter} as POD.}
\vspace{-2mm}
\label{fig:whitebox_eval_clutter}
\end{figure}

\begin{figure}[t]
\centering
\includegraphics[width=0.8\textwidth]{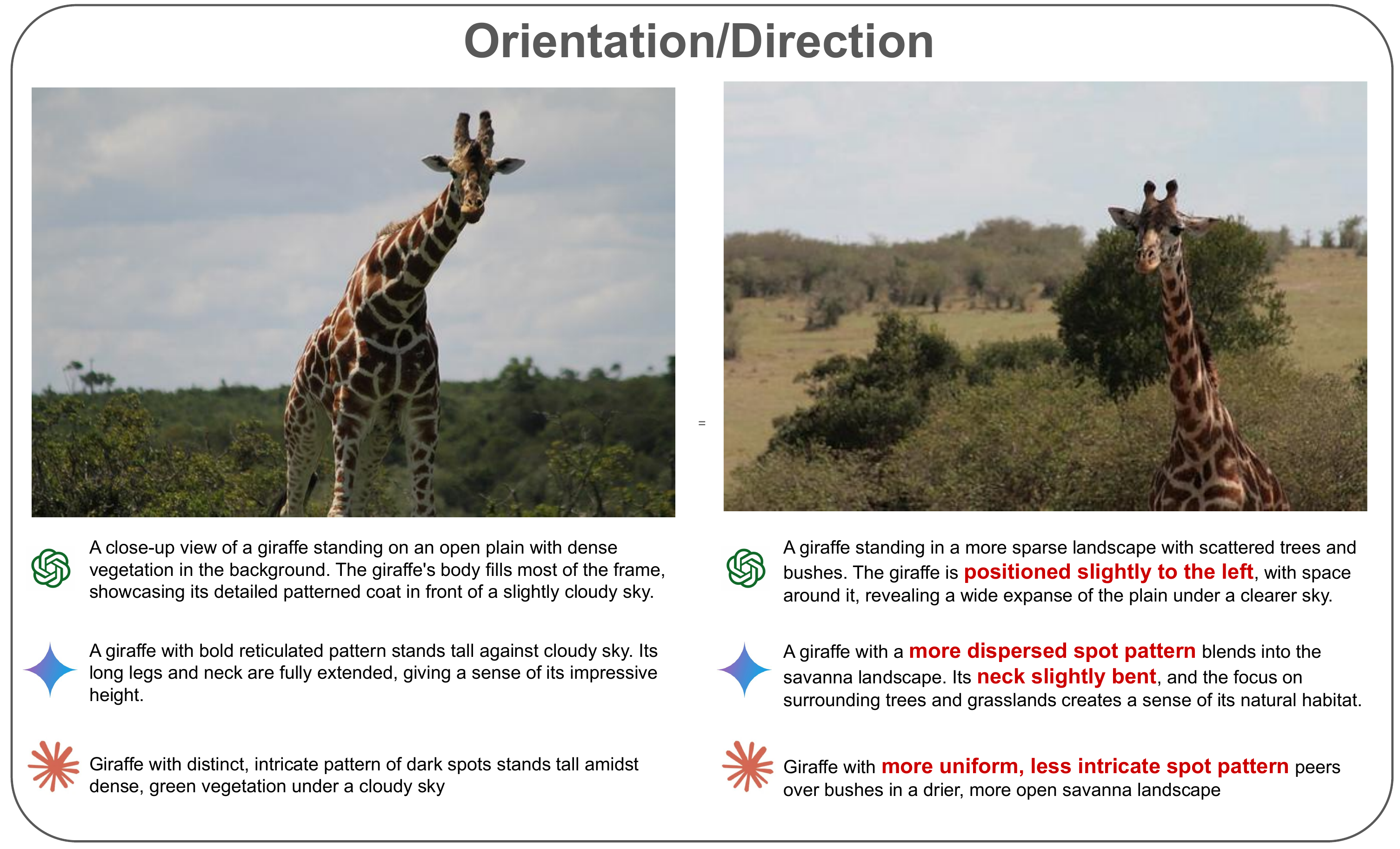}
\vspace{-2mm}
\caption{Outputs of different closed-source MLLMs during \benchmark{} evaluation on an image pair having \textbf{Orientation} as POD.}
\vspace{-2mm}
\label{fig:whitebox_eval_orientation}
\end{figure}

We present qualitative examples in Figs.~\ref{fig:whitebox_eval_state}, \ref{fig:whitebox_eval_positioning}, \ref{fig:whitebox_eval_camera}, \ref{fig:whitebox_eval_scene}, \ref{fig:whitebox_eval_clutter}, \ref{fig:whitebox_eval_orientation}, corresponding to each point of difference.
For each image pair, we also include the captions generated by GPT-4o, Gemini-1.5-Pro, and Claude-Sonnet-3.5.

\end{document}